# C-arm Tomographic Imaging Technique for Nephrolithiasis and Detection of Kidney Stones

Nuhad A. Malalla, Ying Chen *

*Abstract*— in this paper, we investigated a C-arm tomographic technique as a new three dimensional (3D) kidney imaging method for nephrolithiasis and kidney stone detection over view angle less than 180°. Our C-arm tomographic technique provides a series of two dimensional (2D) images with a single scan over 40° view angle. Experimental studies were performed with a kidney phantom that was formed from a pig kidney with two embedded kidney stones. Different reconstruction methods were developed for C-arm tomographic technique to generate 3D kidney information including: point by point back projection (BP), filtered back projection (FBP), simultaneous algebraic reconstruction technique (SART) and maximum likelihood expectation maximization (MLEM). Computer simulation study was also done with simulated 3D spherical object to evaluate the reconstruction results. Preliminary results demonstrated the capability of our C-arm tomographic technique to generate 3D kidney information for kidney stone detection with low exposure of radiation. The kidney stones are visible on reconstructed planes with identifiable shapes and sizes.

## I. INTRODUCTION

KIDNEY stones are small, hard masses that formed inside the kidney when the urine becomes too concentrated with calcium or other minerals. These substances may crystallize and form stones. Kidney stone can move out of kidney to other parts of urinary tract. Nephrolithiasis is the medical term that refers to presence stone in the kidney. Nephrolithiasis can be a problem that cause pain in the abdomen, flank, or groin and other symptoms such as blood in the urine. Kidney stones problem is commonly present in young and middle-aged adults. The number of children (around 5 to 6 years age) getting kidney stone is also rising. In the United States, kidney stone disease affects up to 12% of the American population. In this year, more than 1 million people are expected to have kidney stones. The rate of patients with stones recurrence is about 50% within 5 years and 80% within 20 years [1-12]. To diagnose patients with symptoms of renal or urinary tract disease, various imaging modalities are available such as plain films of the abdomen, renal ultrasonography, intravenous pyelography, and computed tomography. Each technique has their own advantages and disadvantages.

Computed tomography (CT) as a three dimensional x-ray imaging technique is currently the standard x-ray imaging method to confirm the kidney stone detection and to identify the stone location. With CT, different views of the kidney structure were acquired by moving the x-ray source along a circular path around the body. During CT scans, the patient is required to lie on a motorized table that slides into a circular space where the x-ray is taken. CT scans show detailed images of the kidney and also provide detailed cross-sectional images as well as 3D structure of kidney. However, the amounts of radiation delivered by a CT exam is higher that delivered by standard x-ray procedures. To enhance the image quality, CT scans may be done with contrast dye. Contrast dye is a substance taken by mouth or injected into intravenous line, to improve the visibility of particular organ or tissues under study. Consequently, CT scans may introduce risk of allergic reaction to the contrast dye. The contrast dye may cause kidney failure especially in patients with kidney failure or other kidney problems [13-17].

Due to the high amount of radiation during CT procedure, especially for patients who need to track stone migration and fragmentation after undergoing extracorporeal shockwave lithotripsy, many clinical studies in nephrolithiasis and kidney stones detection fields were done on digital tomographic technique for screening patients prior to CT for accessing ureteral and kidney stones [13,15,18,19]. Digital planar tomographic technique provides a dataset of projection image over limited view angle by rotating the x-ray tube around the object to fire a low dosage x-ray beam toward a stationary detector [20-22].

In our study, we investigated a C-arm tomographic technique as a 3D imaging technique for nephrolithiasis and kidney stone detection to reduce radiation dose and examination time [23]. The advantages of using C-arm technique are many, such as the C-arm configuration provides: airgap between the object and detector, a wide range of available projection view angle, a variable magnification and also possibility of nonlinear motion [24].

C-arm tomosynthesis has a C-shaped arm allows both x-ray source and the detector to rotate around the object along a

N. A. Malalla is a PhD candidate in the Department of Electrical and Computer Engineering, Southern Illinois University Carbondale, IL 62901, USA (nuheng@siu.edu).
*Ying Chen is an Associate Professor in the Department of Electrical and Computer Engineering, Southern Illinois University Carbondale, IL 62901, USA.

partial circle scan and to be aligned under each view angle. In this paper, preliminary experiments were done to investigate the C-arm tomosynthesis with low radiation for imaging a kidney phantom. A series of projection images was acquired over 40° angular view. To reconstruct 3D images, four tomographic image reconstruction algorithms were developed including: point by point back projection (BP), filtered back projection (FBP), simultaneous algebraic reconstruction technique (SART) and maximum likelihood expectation maximization (MLEM). Computer simulation study was also done to simulate 3D spherical objects to evaluate the reconstruction results.

## II. METHODS AND MATERIAL

### A. Experimental system

In our kidney experiments, C-arm of the imaging system was rotated around the object to collect the imaging data over limited angular view. Fig. 1 shows the (x, y) plane of C-arm tomosynthesis system which rotates around z-axis with rotation angle β. Both x-ray tube and the detector move along a partial circular orbit that centralized at the rotation center (o).

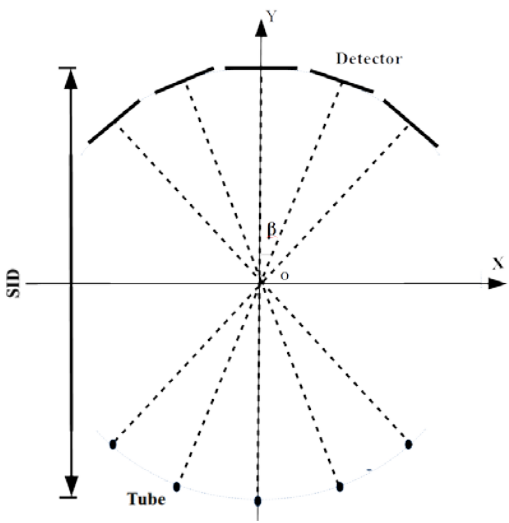

Fig.1 The mid plane (x, y) of C-arm tomosynthesis system

The geometric configuration of our C-arm tomosynthesis system is shown in Fig.2. In the 3D coordinate system, the origin of the 3D coordinate system ($O$) is located at the center of C-arm circular trajectory. The source to image distance (SID) is fixed under any view angle. Tomosynthesis dataset of N projection images was acquired by moving both x-ray source and detector along a partial circular orbit with radius (d) and the center point (O). The half SID is represented by (d). The local coordinate on the detector plane is indicated by (u, v) in which u-axis is perpendicular to z-axis and v-axis is parallel to z-axis. Each line between the sources to the detector passes through the rotation center.

The source is located on the plane (x, y, z=0). The source position $S_p$ is calculated by $Sp = [\, d\cos\beta \; -d\sin\beta \;\; 0]$. The point $D$ is projected onto the detector plane at point $D'$. $D'$ can be find by $[-r\cos(\beta+\alpha) \; r\sin(\beta+\alpha) \; v]$ where $r = \sqrt{(u^2+d^2)}$ and $\alpha = \arctan\left(\frac{u}{d}\right)$.

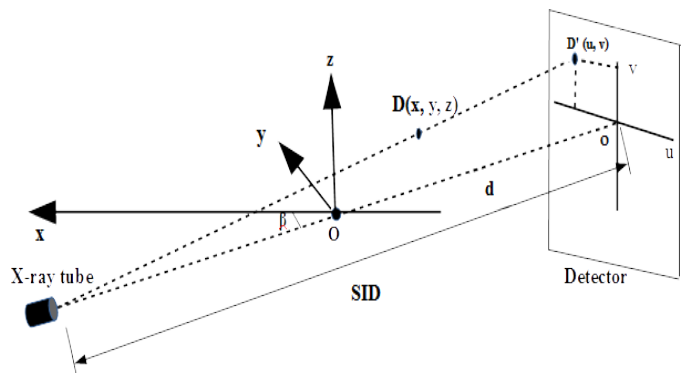

Fig.2 The geometric configuration of our C-arm tomosynthesis system

### B. Reconstruction algorithms

In this paper, two reconstruction methods have been developed to generate the coefficients of system matrix in medical imaging model. Each method differs in their efficiency of calculation accuracy per computing time. Pixel driven method (PDM) is usually used in back projection reconstruction. PDM works with linear interpolation to obtain the pixel value from detector samples. The location on the detector is determined by the intersection between the detector cells and projection lines. Using PDM for projection is rare since PDM generates high frequency artifacts in projection [23].

Point by point back projection (BP) is a mathematic algorithm that works with PDM to reconstruct 3D images. In BP, the pixel value of a given point on reconstructed plane is calculated as the average of obtained projected information of that point by using linear interpolation to obtain the pixel value from detector samples in each projection image over the total number of projection views [25]. The projected location of that point on the detector is determined by the intersection between the detector cells and projection lines. The pixel value of a given point ($\mu_A$) is calculated as

$$\mu = \frac{\sum_{i=1}^{N} P_i}{N}$$

where $P_i$ is pixel value of that point using linear interpolation on $i^{th}$ projection image and $N$ is the total number of projection images [25].

Filtered back projection algorithms (FBP) is a common direct method for reconstructing the collected raw data for a cone beam geometry in a single step [26-30]. The analytical procedure of FBP works on the assumption that the data acquisition are created by transforming the linear attenuation coefficients for the whole imaged volume. For the reconstruction, an analytical inverse transforming can be applied. FBP works on noiseless raw data and ignores Poisson photon noise, detector noise and scattered photon.

FBP procedure starts with normalizing the projection images and filtering the normalizing projection data in the frequency domain before back project the projection data. Several filtering techniques based on the post filtering and de-blurring algorithms for FBP algorithm have been developed to minimize the limited data artifacts, inherent back projection blurry and to improve contrasting features.

Ramp filter is one of high resolution filters that are used with FBP method. The effect of this filter is to suppress low frequencies and raise high frequencies with a linear in between. The high frequency edge of the filter passes the highest frequencies which often do not have useful information and causes noise performance. To filter out these frequencies and balance between resolution and noise, Ramp filter can be combined with several windows functions such as Hanning window. FBP works with PDM to back project the filtered projection images to generate 3D information about the imaged volume.

Advanced developments in computer technologies and mathematical methods overcome limited projection data problem by solving the system with iterative methods. Iterative reconstruction algorithms in this paper are computed with ray driven method (RDM) to calculate the coefficients of the system matrix. Simultaneous algebraic reconstruction technique (SART) [31-33] works on minimizing the error between the measured and the calculated projection data by iteratively updating the unknown linear attenuation coefficients N times per iteration (N is the number of projection images).

The linear attenuation coefficient $\mu$ of $j^{th}$ voxel per iteration $t^{th}$ is expressed as

$$\mu_j^{(t+1)} = \mu_j^t + \lambda \frac{1}{\sum_i w_{ij}} \sum_i \frac{w_{ij}}{\sum_j w_{ij}} \left( A_i - \sum_j w_{ij} \mu_j^t \right) \quad (1)$$

The term $A_i$ is expressed as $A_i = \log\left(\frac{I_i}{I_o}\right)$ where $I_o$ is the detected x-ray intensity and $I_i$ is the incident x-ray intensity. Each element $w_{ij}$ in the system matrix represents the weight of contribution of $j^{th}$ voxel to $i^{th}$ projection ray. The relaxation parameter ($\lambda$) is used to control the update process. The value of $\lambda$ is decreased over limited number of iterations.

Maximum likelihood expectation maximization (MLEM) is developed based on expectation maximization (EM) convex algorithm to maximize the likelihood function [34]. MLEM is under an assumption that the relationship between the incident and transmitted x-ray intensities follows Poisson statistics. Likelihood function is the conditional probability distribution of the detected intensities based on incident intensities and three dimensional attenuation model.

The attenuation coefficient $\mu$ of $j^{th}$ voxel is updated per iteration $t^{th}$ by

$$\mu_j^{(t+1)} = \mu_j^t + \frac{\mu_j^t \sum_i w_{ij} \left( I_i e^{-\sum_i w_{ij} \mu_j^t} - O_i \right)}{\sum_i (w_{ij} \langle w, \mu^t \rangle I_i \, e^{-\sum_i w_{ij} \mu_j^t})} \quad (2)$$

The detected intensity and incident intensity are indicated by $O_i$ and $I_i$ respectively [34].

### C. Preliminary Phantom Study

Preliminary kidney phantom experiments were investigated to generate 3D kidney structure for kidney stone detection with low exposure of radiation. In our preliminary studies, a phantom was formed with two embedded kidney stones inside a pig kidney. The phantom was placed away from the rotation center towards the detector to be visible in all x-ray views during image acquisition. A series of projection images were acquired from 31 angular locations over 40 degrees angular view. The detector size of 153.4 mm x 122.7 mm with 240 μm pixel size was used.

### D. Computer Simulation Study

Computer simulation study was also done to evaluate the performance of C-arm tomosynthesis reconstruction algorithms. 3D spherical object was simulated with imaging parameters of a virtual C-arm tomosynthesis system. The object has 1mm radius and is located at the rotation center of the circular scan. The radius of the circular orbit was 440mm. The parameter SID was 880mm for all projection images. For imaging acquisition, the size of projection images 256 x 256 pixels was used for computational purposes. The pixel size was 240 μm. Dataset of 25 projection images over angular range of 40 degrees was simulated. The performance of each reconstruction algorithm was evaluated by 3D mesh plot of the simulated object on focus reconstruction plane.

### III. RESULTS

### A. Computer Simulation Study Results

The reconstruction results of BP, FBP, MLEM and SART are illustrated in Fig. 3 (a) through (d). In Fig. 3, the 3D mesh plots of reconstructed in-plane normalized pixel intensities of the simulated spherical object are shown specifically. Fig.4 shows the line profiles of the focus reconstructed in-plane of BP, FBP, MLEM and SART. In Fig. 3 and Fig. 4, all investigated algorithms can reconstruct the simulated object by showing sharp in-plane response on the focus reconstruction plane. FBP has edge enhancement performance due to presence Ramp filter as denoted by the arrow. The line profile is calculated as the response along a line passing through the object's center on the in-plane reconstruction slice in spatial domain [29]. Fig. 5 shows the modulation transfer function MTFs for each reconstruction algorithm. MTF[29] is calculated as the frequency response of a line on the in-plane reconstruction slice passing through the center of the spherical object.

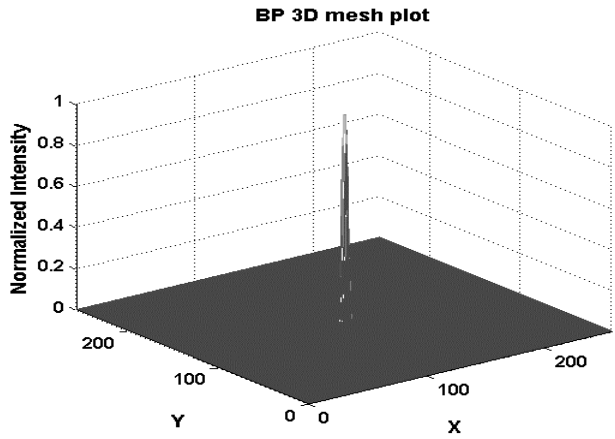

(a)

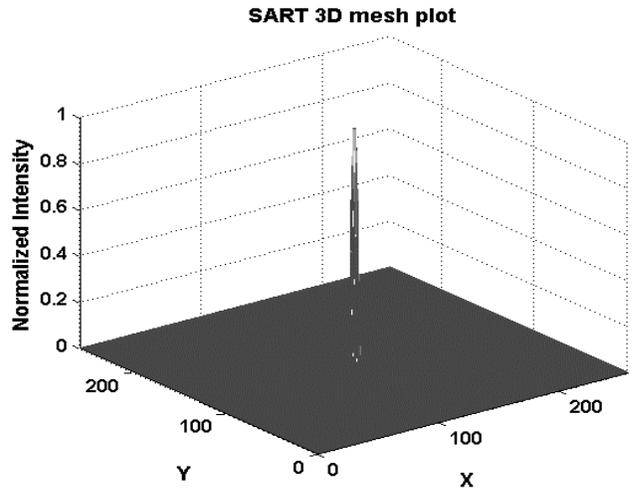

(d)

Fig. 3 Mesh plot of in-plane reconstruction slice passing through the center of simulated spherical object for each reconstruction algorithms: (a) BP, (b) FBP, (c) MLEM and (d) SART.

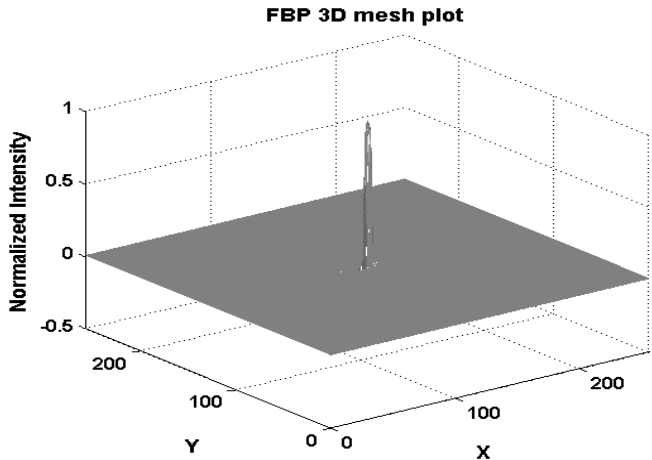

(b)

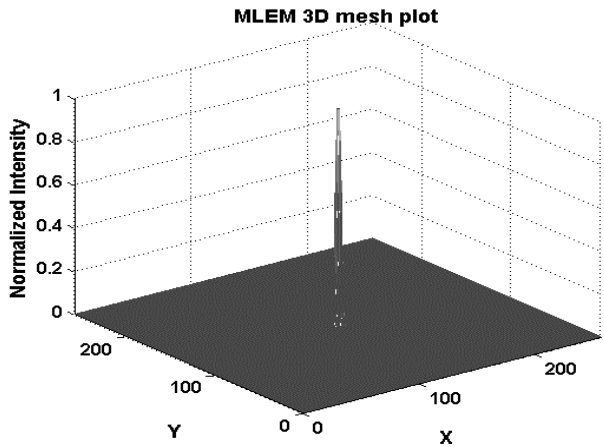

(c)

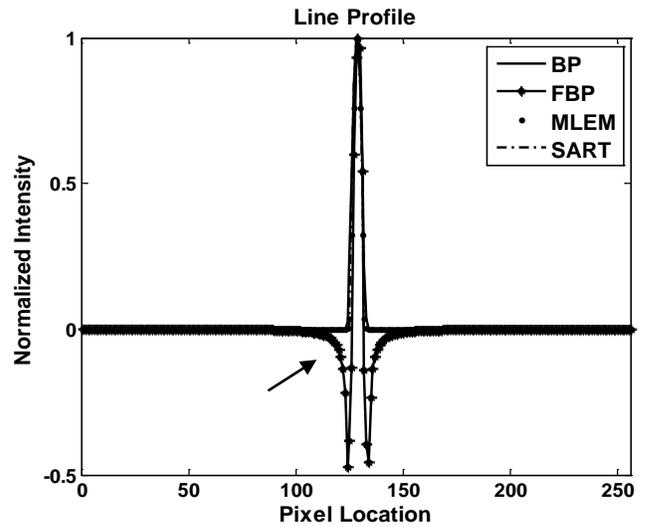

Fig. 4 Line profile through the object center on in-plane slice.

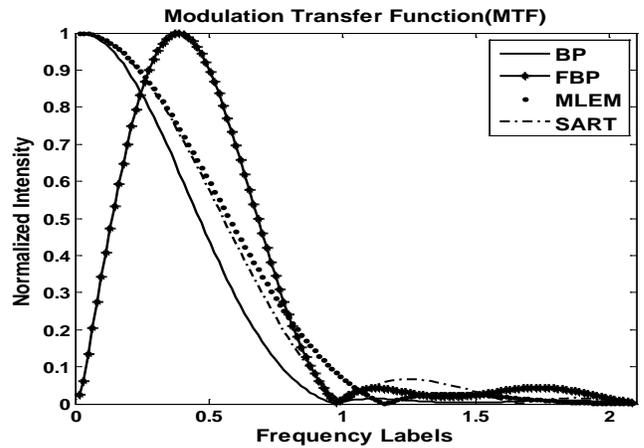

Fig. 5 Modulation transfer function (MTF) through the object center on in-plane slice.

## B. Preliminary Phantom Study Results

Our preliminary results of kidney phantom study are demonstrated in Fig. 6 and 7. Figure 6 shows the in-plane kidney reconstruction slices passing through the kidney stone with BP, FBP, MLEM, and SART reconstruction algorithms specifically.

All investigated reconstruction algorithms are capable to identify embedded kidney stones with 3D positioning information. Fig. 7 (a) through (d) shows the reconstructed region of interests of one embedded kidney stone. Shapes and margins of embedded kidney stone can be identified. Fig. 8 shows the line profiles across the embedded kidney stone for comparison purpose.

Artifact spread function (ASF) [24, 30] plot is shown in Fig.9. FBP, MLEM and SART show better performance in out-of-plane artifacts removal, compared with BP algorithm.

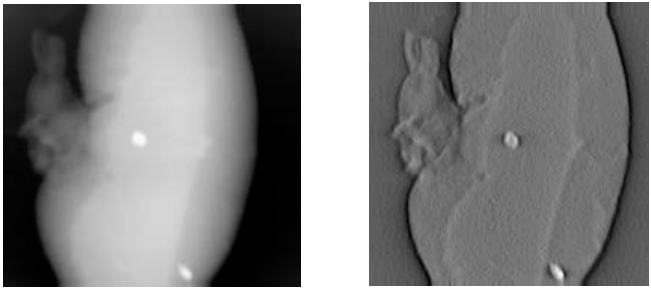

(a)          (b)

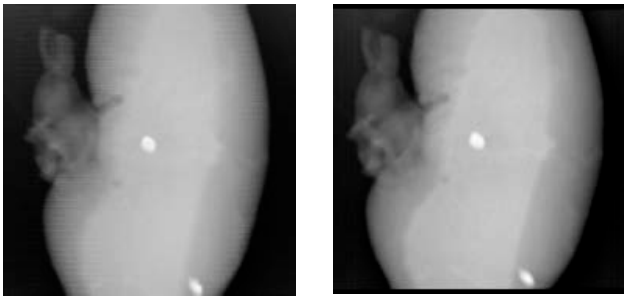

(c)          (d)

Fig. 6 Reconstructed kidney phantom results for each reconstruction algorithms: (a) BP, (b) FBP, (c) MLEM and (d) SART.

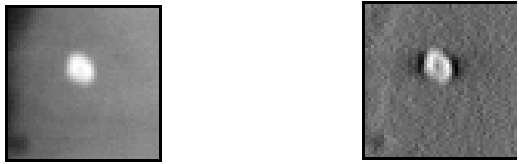

(a)          (b)

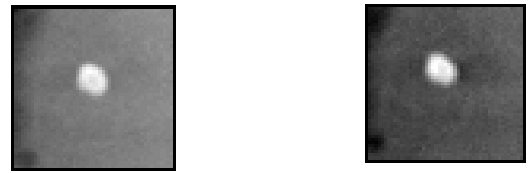

(c)          (d)

Fig. 7 kidney stone results: (a) BP, (b) FBP, (c) MLEM and (d) SART.

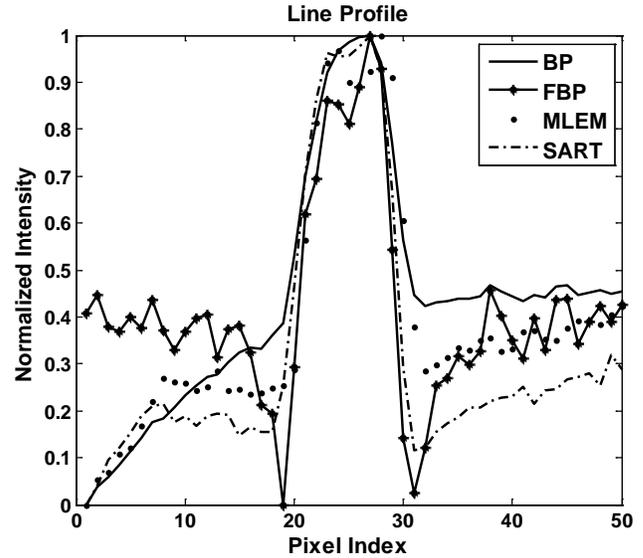

Fig. 8 Line profile through the kidney stone.

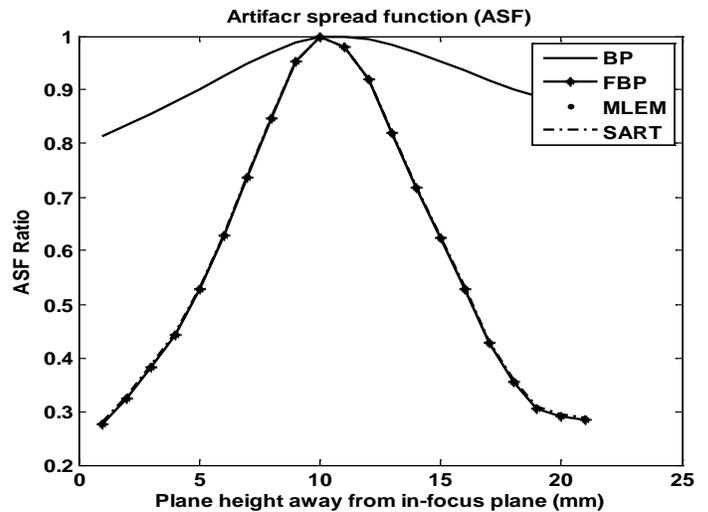

Fig.9 Artifact speared function (ASF) plot.

## IV. CONCLUSION

In this paper, a C-arm tomographic technique is investigated for nephrolithiasis and kidney stones detection. Our proposed technique acquires 2D projection images by rotating C-am gantry around the object over limited angular view with low radiation dosage. Preliminary studies are performed using a kidney phantom with two embedded kidney stones. Four representative reconstruction algorithms have been developed including BP, FBP, SART and MLEM to generate 3D information of the object. Computer simulation studies are done with simulated spherical object to evaluate the performance of each reconstruction algorithm. Preliminary study results demonstrate the capability of C-arm tomographic technique to generate 3D information of kidney structures and to identify the size and location of kidney stones. Compared with other reconstruction algorithms, BP shows more blurry artifacts and less out-of-plane structure removal. C-arm tomographic technique shows capability to provide 3D information with low dosage of radiation. Further study will be done to investigate and optimize our kidney tomosynthesis with other phantoms and cadaver experiments.